\documentclass[letterpaper, 10 pt, conference]{ieeeconf}  
\usepackage{graphicx}

\IEEEoverridecommandlockouts                              

\title{\LARGE \bf
LAMP: Long-Horizon Adaptive Manipulation Planning \\ for Multi-Robot Collaboration in Cluttered Space
}

\author{$\text{Shuai Zhou}^1, \text{Yorai Shaoul}^1 \text{ and Jiaoyang Li}^1$
\thanks{$^{1}$Authors are with the Robotics Institute, Carnegie Mellon University, Pittsburgh, PA 15213, USA. shuaizho@cs.cmu.edu}%
\thanks{Project page: \url{https://multi-robot-lamp.github.io/}}
}

\usepackage{multirow}
\usepackage{booktabs}
\usepackage{subcaption}
\usepackage{pifont}
\usepackage{hyperref}
\hypersetup{hidelinks}
\usepackage[dvipsnames]{xcolor}
\usepackage{graphicx}
\usepackage{float}
\usepackage{amsmath}
\usepackage{amsfonts}
\usepackage{amssymb}
\usepackage{bbm}
\usepackage{cite}
\usepackage{caption}

\usepackage{algorithm}
\usepackage[noend]{algpseudocode}

\algdef{SE}[FUNCTION]{Function}{EndFunction}[2]{\textbf{function} \textsc{#1}(#2)}{\textbf{end function}}
\algtext*{EndFunction}

\begin{document}

\maketitle
\thispagestyle{empty}
\pagestyle{empty}

\begin{abstract}
	Multi-robot manipulation requires jointly reasoning about contact formations, robot motions under coupled dynamics, and collision avoidance. Systematically searching over this large space is difficult and becomes increasingly intractable as the number of robots grows, the task horizon lengthens, or the scene becomes more densely cluttered. Existing approaches therefore either learn to solve the problem end-to-end via reinforcement learning or restrict planning to a simpler surrogate problem, such as planning object motions while learning short-horizon contact primitives. However, neither paradigm scales to the problem instances we target: long-horizon multi-robot manipulation in extremely dense environments. In this paper, we propose Long-horizon Adaptive Manipulation Planning (LAMP), a framework combining a generative model for manipulation with classical planning for long-horizon reasoning. We instantiate our framework with two algorithms leveraging insights from established planning techniques, A* and lazy search: LAMP-A*, which systematically searches over the coupled object-robot space, and LAMP-Lazy, a lazy planner that enables real-time replanning through deferred evaluation. Experiments in challenging simulated environments demonstrate that our approach solves complex long-horizon tasks in highly cluttered environments that prior methods cannot handle.
\end{abstract}


\section{Introduction}
\label{sec:introduction}
	Collaborative multi-robot manipulation is fundamental to enabling robots to handle tasks that exceed the capabilities of a single robot. By coordinating their physical interactions, multiple robots can jointly move and position objects in complex environments, a capability critical to applications ranging from warehouse logistics to construction. This collaborative approach unlocks manipulation tasks involving larger, heavier, or more awkwardly shaped objects that would otherwise be infeasible for a single robot.

Unlike single-robot manipulation, multi-robot manipulation poses unique challenges: Robots must reason about the coupled dynamics of objects under contact while dynamically allocating and reallocating interactions among team members. The resulting search space that spans robot-object assignments, contact formations, and coordinated motions grows combinatorially with the number of robots, making planning computationally prohibitive.

\begin{figure}[t]
\vspace{0.25cm}
\centering
\includegraphics[width=0.6\columnwidth]{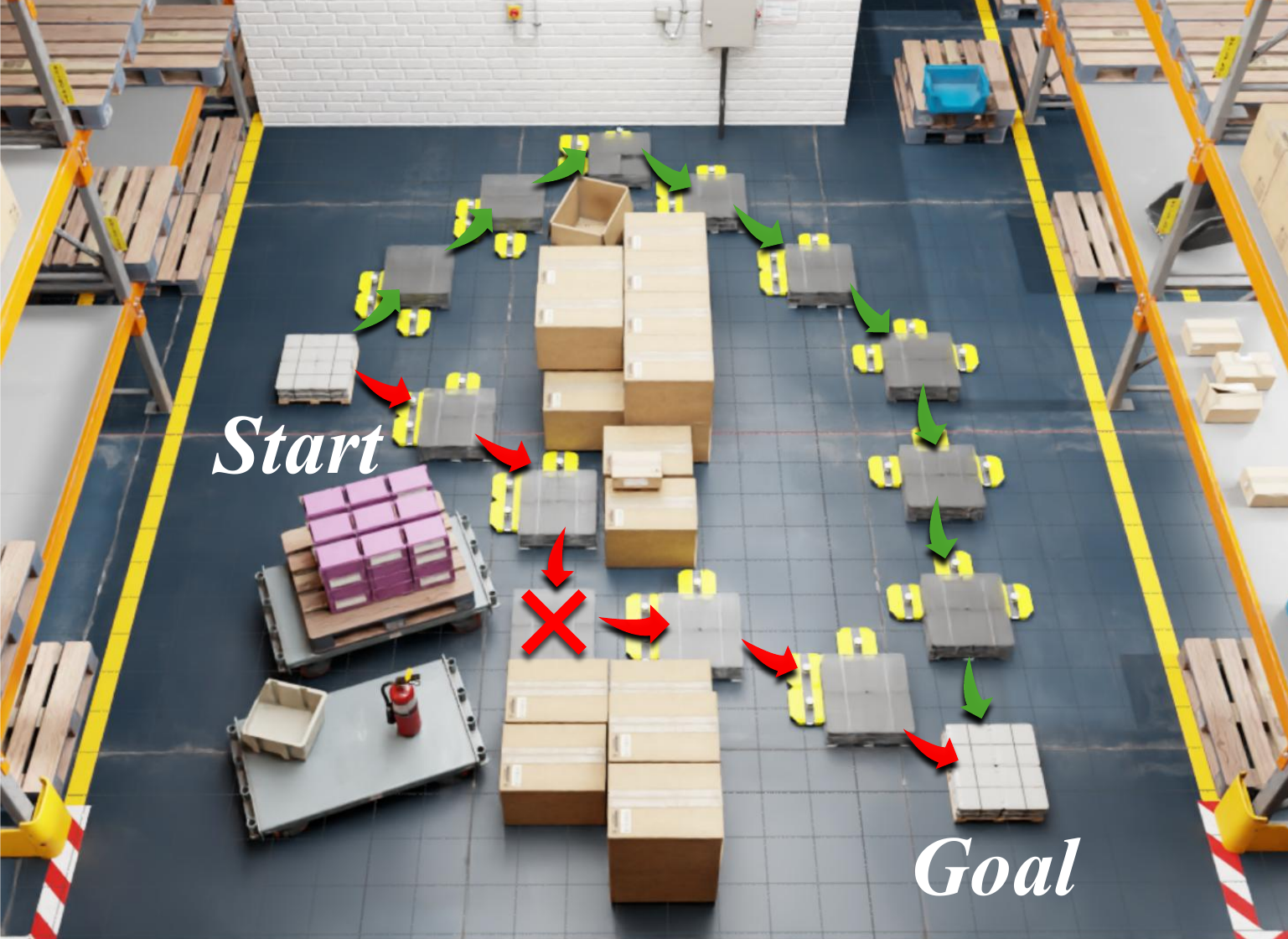}
\caption{Motivating example in a cluttered warehouse environment where robot manipulation feasibility must be integrated into object path planning. The yellow ground robots must collaboratively transport a pallet cargo from start to goal, navigating around traffic barriers and stationary cargo acting as obstacles. Two candidate paths for the object are illustrated with arrows: the red path is shorter but becomes infeasible at the marked location ($\times$) where insufficient space prevents robots from establishing manipulation contacts; the green path ensures feasibility throughout despite higher cost.}
\label{fig:idea}
\vspace{-0.6cm}
\end{figure}

Existing approaches address this challenge from complementary perspectives. Pure learning-based methods~\cite{fyuming2025MAPush} capture manipulation behaviors directly from data but often sacrifice generalization and scalability. Pure planning-based methods~\cite{tang2024collaborative} leverage structure to compute coordinated trajectories, but require accurate system dynamics and are brittle under real-world disturbances.
Recent work~\cite{shaoul2025Gco} combines planning for long-horizon reasoning with learning-based generation of local manipulation motions. However, it requires object trajectories to provide sufficient clearance for robots to make contact from any position around the object, limiting its applicability in confined spaces.
Fig.~\ref{fig:idea} illustrates a representative failure case: two candidate paths exist for transporting the object from start to goal. The red path appears shorter and would be prioritized by prior methods, but fails to provide sufficient clearance for robots to establish manipulation contacts at critical intermediate waypoints (marked by a red ×), rendering execution infeasible. The green path, on the other hand, allows for robots to successfully manipulate the object towards its goal despite the higher path cost. 
This scenario highlights a key insight: instead of solving a surrogate object-planning problem and applying learned manipulation models only after the fact, we should incorporate robot-level manipulation feasibility directly into the search. By using learned predictions during path planning, not merely as a post hoc filter, our approach reasons about feasibility where it matters most. Moreover, it is designed for closed-loop execution, enabling fast replanning to recover from drift and imperfect predictions rather than assuming clean execution. As a result, the method remains robust in cluttered environments and generalizes to new obstacle layouts and task horizons without retraining.

In this paper, we present a unified \textbf{Long-horizon Adaptive Manipulation Planning (LAMP)} framework for Multi-Robot collaboration with two key contributions: \textbf{(i)} \textbf{LAMP-A*} that leverages a 
learned manipulation model to systematically search the joint space of both object and robot configurations, contact assignments, and generate executable long-horizon manipulation plans. \textbf{(ii)} \textbf{LAMP-Lazy} that dramatically reduces planning time through lazy evaluation and penalization, allowing us to rapidly replan when execution deviates from the planned trajectory due to inaccurate model predictions. We demonstrate that our framework solves challenging tasks beyond the reach of previous work with improved robustness, and reduced planning time compared to our LAMP-A* baseline, all without model retraining.

\begin{figure*}[t]
\vspace{0.18cm}
\centering
\includegraphics[width=0.7\textwidth]{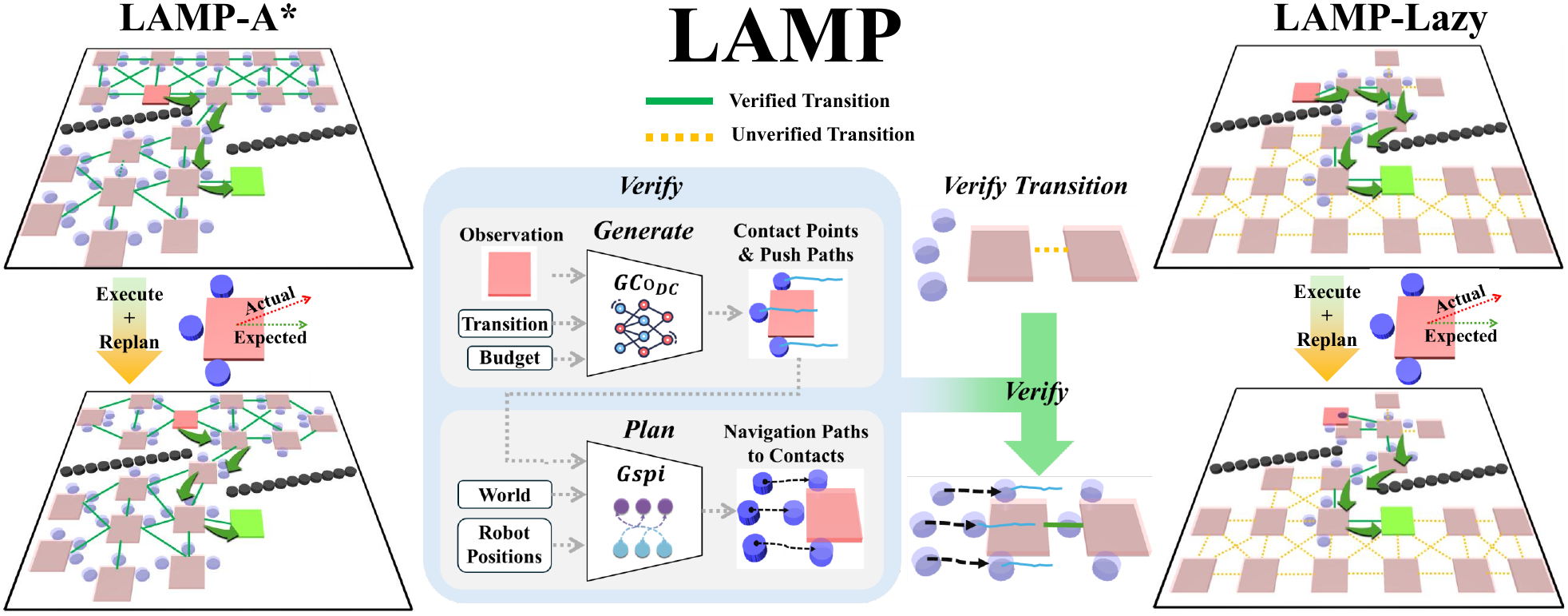}
\caption{Overview of LAMP framework: Green arrows show the planned path to the goal. \textbf{Left (LAMP-A*):} Performs A* search over coupled object-robot configurations. Green edges represent verified transitions coupling object motions with feasible robot manipulation trajectories. When execution drift occurs, it replans from the new configuration. \textbf{Right (LAMP-Lazy):} Uses D* Lite to generate object-centric paths (orange edges), then verifies each transition by generating corresponding robot manipulation trajectories (producing green edges). When drift occurs, D* Lite incrementally computes a new path, and cached verified transitions enable rapid replanning. \textbf{Center (Verify):} Given a desired object transition, invokes $\text{GCo}_{DC}$ to generate robot manipulation trajectories and contact points, then uses Gspi to plan navigation paths that move robots from their current positions to these contacts. (Note that discretization shown is coarser than in experiments for clarity.)}
\label{fig:overview}
\vspace{-0.6cm}
\end{figure*}

\section{Background}
\label{sec:related_work}
	This section introduces the problem setting and positions our work within prior literature. We first formalize the multi-robot non-prehensile manipulation problem, then review related work on multi-robot manipulation and on integrating planning with learning. We finally discuss Generative Collaboration (GCo)~\cite{shaoul2025Gco} in greater detail, as it is both a particularly relevant prior method and a baseline for our approach.

\subsection{Problem Formulation}

We consider multi-robot non-prehensile manipulation in a shared planar workspace with static obstacles. The objective is to coordinate a team of robots to transport an object to a target configuration while maintaining collision-free motion and adapting online to execution uncertainty.

Let $\mathcal{R} = \{R^1, \dots, R^N\}$ denote a team of $N$ robots operating in a shared workspace $\mathcal{W} \subseteq \mathbb{R}^2$. Each robot $R^i$ is modeled as a disk of radius $r$, with configuration $q_R^i \in \mathcal{Q}_R \subseteq \mathbb{R}^2$. A robot trajectory $\tau^i$ is a sequence of configurations over time. We consider a movable object $O$ with configuration $q_O \in \mathcal{Q}_O \subseteq SE(2)$, together with static obstacles that neither the robots nor the object may intersect.
The task is to transport $O$ from an initial configuration to a goal configuration through non-prehensile interactions. We seek coordinated robot trajectories that realize the desired object motion while avoiding robot-robot, robot-obstacle, and object-obstacle collisions. Due to uncertain contact dynamics and execution disturbances, robot trajectories $\mathcal{T} = \{\tau^1, \ldots, \tau^N\}$ must be calculated online and dynamically replanned in response to observed object drift from planned positions, enabling adaptive replanning and reassignment of robot-object interactions as needed.

\subsection{Related Work}

Prior work relevant to our setting falls into two broad categories. The first addresses multi-robot manipulation directly, through either planning-based or learning-based methods. The second studies how learned components can be integrated into structured planning frameworks. 

Multi-robot manipulation has been studied under a broad range of assumptions and problem formulations. Existing methods differ primarily in how explicitly they model robot-object interaction, how much task structure they assume, and how well they handle long-horizon execution in cluttered environments.
Early planning-centric methods typically rely on strong structural assumptions, such as predefined manipulation primitives or simplified interaction models. For instance, \cite{pan2021general} proposes a general framework for coordinating multiple manipulators, but it depends heavily on hand-designed primitives, which can limit adaptability. Likewise, \cite{hartmann2022long} studies large-scale multi-robot rearrangement and achieves strong scalability, but does not model robot-object physical interactions. A separate line of work learns collaborative manipulation policies end-to-end~\cite{fyuming2025MAPush, nachum2019multi, xiong2024mqe}. These methods can capture behaviors that are difficult to engineer manually, but they often become brittle in cluttered environments or over long horizons, where compounding errors may drive execution into out-of-distribution areas. 

A natural alternative is to combine planning and learning. Broadly speaking, planning provides global search and consistency, while learning supplies local models or decision rules that are difficult to specify analytically. In task and motion planning~\cite{garrett2021integrated}, prior work has used learning to model actions~\cite{liu2025one, mao2023learning}, motion shortcuts~\cite{liu2025slap}, operators~\cite{toms2023learnSkill, Rohan2022learntran}, and predicates~\cite{Li2025RSS, silver2023predicate}. In multi-robot planning, learning has likewise been used to provide heuristic guidance~\cite{HuangAAAI22} or local coordination strategies~\cite{JiangICRA25}. Most relevant to us, in multi-robot manipulation, GCo \cite{shaoul2025Gco} interleaves object-level planning with manipulation policy inference to outperform pure-planning and pure-learning approaches.

However, many hybrid approaches still leave an important gap: they either invoke learned models only after the planning stage, or they rely on learned outputs without strong mechanisms for detecting and recovering from inaccuracies during execution. This limitation is especially pronounced in multi-robot manipulation, where local feasibility depends on tight geometric constraints and uncertain contact dynamics. In this work, we draw inspiration from aspects of GCo and therefore provide a closer discussion of GCo below.

\subsection{Generative Collaboration (GCo)} \label{bg:gco}

GCo combines object-centric planning with learned short-horizon manipulation generation. At a high level, it plans a path for the object, selects a local object motion, and then generates robot behaviors intended to realize that motion. An anonymous multi-robot motion planner is subsequently used to route robots to the required contact locations before execution. Repeating this process allows the object to be advanced incrementally toward the goal.

At the core of this approach are two key components. First, a \textit{manipulation model} that takes as input an object observation $\mathcal{I}$, a desired short-horizon object motion primitive $m_O$, and a robot budget $B$, and generates a set of contact points $\mathcal{C}_{manip} = \{ c_i \in \mathbb{R}^2 \}_{i=1}^{B}$ together with corresponding pushing trajectories $\mathcal{T}_{manip} = \{ \tau_i^{manip} \}_{i=1}^{B}$. Second, an \textit{anonymous multi-robot motion planner} (AMRMP) that takes current robot configurations $s_R = (q_R^1, \dots, q_R^N)$ and a set of target contact points $\mathcal{C}_{manip}$, and jointly computes robot-contact assignments and collision-free trajectories to convey robots to those contact points, denoted $\mathcal{T}_{move}$.

This approach is highly scalable: inference is quick, and high-level object-centric planning in a simplified surrogate space is tractable. However, this abstraction assumes that the selected object motions can be realized by feasible robot motions. In confined environments, this assumption can fail, since a valid object path may require robot maneuvers that are blocked by surrounding geometry. This limitation motivates the central idea of our approach. Rather than planning over object motions alone and invoking learned manipulation only afterward, we incorporate robot-level manipulation feasibility directly into the search. As in GCo, we use learned short-horizon manipulation proposals and anonymous multi-robot routing as modular components. Unlike GCo, however, we tightly couple these components in a planning framework that validates proposed motions during search and can rapidly replan in response to execution drift. Because this adaptation occurs online, our method generalizes to new obstacle layouts and task horizons without retraining.

\section{Methodology}
\label{sec:methodology}
	We introduce Long-horizon Adaptive Manipulation Planning (LAMP), a framework for multi-robot collaborative manipulation in cluttered environments. The core idea is to integrate data-driven robot-level manipulation feasibility verification directly into object-level kinematic path planning. This interleaved approach validates that each object transition can be realized through feasible robot manipulation, both in terms of robot-object interactions as well as collision avoidance. An overview is shown in Fig.~\ref{fig:overview}. LAMP does so by orchestrating two modular sub-components, a short-horizon manipulation interaction generator and an anonymous multi-robot motion planner, within a unified planning framework. In practice, we make use of $\text{GCo}_{DC}$ and Gspi from GCo~\cite{shaoul2025Gco}, but do not rely on their internal structure and can use any sub-component implementation in our framework.

A central design question is when and how to invoke feasibility verification during object-level search. A natural approach is to verify each transition eagerly as it is considered. We instantiate this idea in \textbf{LAMP-A*}, which, at its core, performs an A* search over the coupled object-robot configuration space, where transitions are generated using the learned manipulation model and multi-robot motion planner. While LAMP-A* can solve extremely constrained problems, it explores the large search space inefficiently and is ill suited to recover from execution deviations. Thus, leveraging insights from lazy planning, we present \textbf{LAMP-Lazy}: an algorithm that defers manipulability validation until a complete object path is obtained and maintains an incremental evaluation tree to efficiently reuse prior results during replanning, enabling real-time closed-loop execution.
\subsection{LAMP-A*}

An intuitive and principled approach for computing long-horizon collaborative manipulation plans
is to systematically search the space of coupled robot-object configurations. One way to represent this space is in a graph, where each vertex represents a joint object-robot configuration, and edges correspond to motions for both objects and robots. Since objects unfortunately cannot move on their own, any object movement along an edge must be induced by the corresponding robot motions. 
Access to this graph would naturally allow using algorithms like A* for finding robot motions that convey an object to its goal. However, constructing this graph is hard: each edge requires robot trajectories that realize the desired object motion, yet without accurate dynamics models, we cannot determine a priori which robot actions will successfully produce the intended object movements.

LAMP-A* addresses this challenge by constructing the graph implicitly during the search procedure rather than a priori. The key idea is to verify edge validity on demand: for each proposed object transition, we invoke learned manipulation and motion planning models that together determine whether robots can reach appropriate configurations and successfully execute the transition to realize the object motion (see Fig.~\ref{fig:overview}). This allows us to reason over the complete coupled configuration space without constructing an explicit graph a priori. Running an A* search with this on-demand verification yields a full-horizon manipulation plan in which every transition has been validated before execution. When we find a configuration in which the object is at the goal, a simple backtracking retrieves the complete manipulation plan by concatenating the trajectories along the path.

\begin{algorithm}[tb]
\vspace{0mm}
    \small
	\caption{$\rm \text{LAMP-}A^\star $}
\begin{algorithmic}[1]
\Require $\text{object } O, \text{robot }R, \text{policy } \text{GCo}_{DC}, \text{planner }\text{Gspi}$
\State $s_{init} \leftarrow \textsc{State}(q_O^{init}, q_R^{init}, \emptyset)$
\State $Open.add(s_{init})$
\While {$\text{time limit is not reached}$}
\State $s_{curr} \leftarrow Open.pop()$ \label{astar:openpop}
\State $q_O^{curr} \leftarrow s_{curr}.q_O;q_R^{curr} \leftarrow s_{curr}.q_R$
\State $\textbf{if } \textsc{ReachGoal}(s_{curr}) {\rm \textbf{ then return }} \textsc{BackTrack}(s_{curr})$\label{astar:backtrack}
\For{each motion primitive $m_O$} \label{astar:expand state}
\State $\mathcal{T}_{move},\mathcal{T}_{manip}, \mathcal{C}_{R} \leftarrow \textsc{Verify} (q_O^{curr},m_O,q_R^{curr})$ \label{astar:verify}
\State $\textbf{if } \mathcal{C}_{R} = \emptyset  {\rm \textbf{ then continue }}$
\State $q_R^{next} \leftarrow \mathcal{T}_{manip}.last();\mathcal{T}_R^{(s_{curr}, s)} \leftarrow \mathcal{T}_{move} \oplus \mathcal{T}_{manip}$
\State $q_O^{next} \leftarrow q_O^{curr} * m_O$ \Comment{Pose after motion primitive}
\State $s \leftarrow \textsc{State}(q_O^{next},q_R^{next}, s_{curr})$ \label{astar:candidate}
\State $s_{prev} \leftarrow \textsc{Explored}(q_O^{next},q_R^{next})$\label{astar:prev state}
\State $\textbf{if } s_{prev} \neq \emptyset  {\rm \textbf{ then }} \textsc{UpdateCost}(s,s_{curr}, s_{prev})$
\State $ {\rm \textbf{else }\textsc{AddChild}(s_{curr}, s); Open.add(s)}$
\EndFor
\State $Open.re\text{-}add(s_{curr})$ \Comment{With Penalty} \label{astar:readdstate}
\EndWhile
\Function{Verify}{$q_O^{curr},m_O,q_R^{curr}$}
\State $\mathcal{I} \leftarrow \textsc{Observe}(q_O^{curr})$ \label{astar:observe}
\For{each budget $B$ in range $N$ to 1} \label{astar:budget} 
\For{$\mathcal{C}_{R}, \mathcal{T}_{manip} \in \text{GCo}_{DC}(\mathcal{I}, m_O, B, K)$} \label{astar:GCO}
\State $\textbf{if } \textsc{Collide}(\mathcal{C}_{R}, \mathcal{T}_{manip})  {\rm \textbf{ then continue }}$\label{astar:collide}
\State $\textsc{Reassign}(\mathcal{C}_{R}, \mathcal{T}_{manip},q_R^{curr})$ \label{astar:reassign}
\State $\mathcal{T}_{move} \leftarrow \text{Gspi}(q_R^{curr},C_{R})$\label{astar:Gspi}
\State $\textbf{if } \mathcal{T}_{move} \neq \emptyset  {\rm \textbf{ then}}$ \Return $\mathcal{T}_{move}, \mathcal{T}_{manip}, \mathcal{C}_{R}$ \label{astar:return_verify}
\EndFor
\EndFor
\State $\Return \ \emptyset, \emptyset, \emptyset$\label{astar:failure}
\EndFunction
\Function{UpdateCost}{$s, s_{curr}, s_{prev}$}
\State $\textsc{AddChild}(s_{curr}, s_{prev}$) \label{astar:connect}
\State $\textbf{if } s.cost \geq s_{prev}.cost  {\rm \textbf{ then return}}$
\State $s_{prev}.cost \gets s.cost; s_{prev}.parent \gets s_{curr}$
\State $Open_{update} \leftarrow \{s_{prev}\}$
\While{$Open_{update} \neq \emptyset$} \label{astar:bfs begin}
\State $s_{top} \leftarrow Open_{update}.pop()$ 
\For{$s_{child} \in s_{top}.children$}
\If{$s_{child}.cost > s_{top}.cost + \mathcal{T}_R^{(s_{top}, s_{child})}.cost$}
\State $s_{child}.cost \leftarrow s_{top}.cost + \mathcal{T}_R^{(s_{top}, s_{child})}.cost$
\State $s_{child}.parent \gets s_{top}$
\State $Open_{update}.add(s_{child})$ \label{astar:bfsend}
\EndIf
\EndFor
\EndWhile
\EndFunction
\end{algorithmic}
\label{alg:Astar}
\vspace{-0.1cm}
\end{algorithm}

Formally, we define a search state as $s = (q_O, q_R, s_{parent})$, where $q_O$ is the object configuration, $q_R$ denotes the robot configurations, and $s_{parent}$ is the parent state for backtracking. An edge is associated with robot trajectories $\mathcal{T}_R^{(s, s')}$ that manipulate the object from the object configuration in state $s$ to that in state $s'$.

LAMP-A* proceeds as follows (Alg.~\ref{alg:Astar}). We begin from the initial configuration of the robots and the object, and iteratively pop search states from the priority queue (line~\ref{astar:openpop}) to expand the reachable space. If the object reaches the target configuration (line~\ref{astar:backtrack}), we backtrack through the solution path and concatenate the robot trajectories associated with each edge to obtain the full-horizon manipulation plan. Otherwise, we expand the current search state (line~\ref{astar:expand state}) by evaluating a set of predefined object motion primitives one by one. For each primitive $m_O$, we invoke $\textsc{Verify}$ (line~\ref{astar:verify}) to determine whether feasible manipulation trajectories can be generated; details of $\textsc{Verify}$ are given in Sec.~\ref{method:astar:verify}. If successful, $\textsc{Verify}$ returns trajectories $\mathcal{T}_{move}$, $\mathcal{T}_{manip}$ and contacts $\mathcal{C}_{R}$, which we use to generate a successor state $s$ (line~\ref{astar:candidate}) and construct an edge from the current state to $s$ associated with $\mathcal{T}_R^{(s_{curr}, s)} = \mathcal{T}_{move} \oplus \mathcal{T}_{manip}$. Before inserting $s$ into the open list, we check whether an existing state has already been reached at this joint object-robot configuration (line~\ref{astar:prev state}). If so, we connect the current state to that existing state and invoke $\textsc{UpdateCost}$ to determine whether a shorter path has been found (details in Sec.~\ref{method:astar:updatecost}). Otherwise, we add the new state to the priority queue for future expansion. 
Before moving to the next iteration, we re-insert the current search state back into the queue with an added penalty cost ($10^4$ in our implementation) (line \ref{astar:readdstate}) to allow re-expansion with a new set of successors. We explain this in Sec.~\ref{method:astar:verify}. 

We now discuss the subroutines of LAMP.
\subsubsection{Verify} \label{method:astar:verify} 
Given an object transition from $q_O^{curr}$ to $q_O^{next}$, the aim of this function is to generate robot trajectories that achieve it. This is done by using both the manipulation policy and AMRMP sub-components. The function begins by obtaining an observation of the object, in our case a binary image $\mathcal{I}$ (line \ref{astar:observe}), and batch generates a set of $K$ candidate contact points and manipulation trajectories, 
in our case by $\text{GCo}_{DC}$, for each possible budget (lines~\ref{astar:budget}--\ref{astar:GCO}); details of robot budgets are discussed in Sec.~\ref{method:astar:robobudget}.
We then iterate through each candidate pair $(\mathcal{C}_R, \mathcal{T}_{manip})$: for each candidate, we first check for robot-obstacle collisions and filter invalid proposals (line~\ref{astar:collide}). For the remaining trajectories, we invoke $\textsc{Reassign}$ to pair robots with contact points based on their shortest distance. Unpaired robots are assigned to collision-free waiting locations\footnote{We find safe waiting locations for a robot via a breadth-first search from its current position until a collision-free configuration is found, ensuring that waiting locations remain close to the robots' current positions.} (line \ref{astar:reassign}). When the number of contact points matches the number of robots, we invoke Gspi to perform anonymous multi-robot motion planning to convey the robots to their assigned contact configurations (line~\ref{astar:Gspi}).
If Gspi successfully computes collision-free paths, the transition is validated and we return the corresponding trajectories $\mathcal{T}_{move}$, $\mathcal{T}_{manip}$ and the contact points $\mathcal{C}_{R}$  (line \ref{astar:return_verify}). If all $K$ candidates fail validation, the function returns failure (line~\ref{astar:failure}). However, this does not necessarily mean that the object transition itself is infeasible. Due to the stochastic nature of the learned model, a different sampling in a future call may yield valid manipulation trajectories. To preserve this possibility while avoiding excessive recomputation, we re-insert the parent search state into the priority queue with an added penalty cost (line \ref{astar:readdstate}), allowing the transition to be re-explored later if more promising paths are exhausted.

\subsubsection{Update Cost}
\label{method:astar:updatecost}
When a new edge to an existing state $s_{prev}$ is built and a shorter path to it is found, we must update its cost and propagate this change to its successors to ensure we always obtain the shortest path. Specifically, we first establish an edge between the existing search state $s_{prev}$  and the current state $s$ (line \ref{astar:connect}). If this could lead to a shorter path from the start state to $s_{prev}$, we propagate this information to its successors and determine whether any state needs to reset its parent by performing a best-first search ordered by ascending state cost (lines \ref{astar:bfs begin}--\ref{astar:bfsend}). We iteratively update each successor's cost if the new cost is smaller, and reset its parent until no more updates need to be made. 

\subsubsection{Robot Budget}\label{method:astar:robobudget}
The robot budget for short-horizon manipulation presents a fundamental trade-off: larger budgets yield more accurate manipulation trajectories~\cite{shaoul2025Gco} but require more workspace clearance, which may not be available in cluttered environments; smaller budgets are easier to validate but less accurate, leading to increased execution drift. To prioritize manipulation accuracy while maintaining robustness, we iterate over budgets in descending order (line \ref{astar:budget}), beginning with the maximum number of robots and progressively reducing until a feasible solution is found. This strategy enables high-precision manipulation in open spaces while gracefully degrading to lower budgets when necessary to ensure feasibility in cluttered configurations. 

By systematically exploring the object-robot joint configuration space with manipulation feasibility checking, LAMP-A* can find long-horizon manipulation plans. However, it suffers from two critical limitations. First, it is brittle to execution divergence: the learned manipulation model may generate trajectories that, due to approximation errors in modeling complex contact dynamics, cause the object to drift from its planned position during execution. One remedy is to actively replan when execution deviates from the plan. However, this leads to the second limitation: LAMP-A* is computationally expensive due to its large branching factor, making real-time replanning infeasible. The computational bottleneck lies primarily in the expensive calls to the learned model for trajectory generation and to the AMRMP planner for robot approach and assignment. 
We address these limitations in our main algorithm LAMP-Lazy.

\subsection{LAMP-Lazy}

To circumvent the limitations of LAMP-A* while still reasoning over the complete search space of coupled robot-object configurations, we draw on lazy search algorithms to reduce the burden of evaluating transition feasibility while also making our planner tractable and suitable for closed-loop execution, being efficient enough to solve new planning problems frequently and correcting for drift.

Classical lazy planning reduces computation by postponing expensive edge evaluations until they are needed for a candidate solution path. Lazy PRM~\cite{Robert2000LPRM} applies this idea to sampling-based roadmaps by delaying collision checking on roadmap edges, while LazySP~\cite{dellin2016unifying} generalizes it to shortest-path search on graphs whose edge costs or validities are expensive to evaluate. Lazy evaluation reduces verification cost, but closed-loop execution also requires fast replanning under object and robot execution drift. D* Lite~\cite{skoenig2002dstarlite} is well-suited for these purposes. D* Lite is an incremental search algorithm that repairs plans when the environment changes rather than replanning from scratch, which significantly reduces computational cost. We therefore use it as the underlying object-level planner. It can handle dynamic environments where edge costs change (e.g., obstacles appear or edges become infeasible) and where the start configuration shifts over time. Specifically, D* Lite operates through three conceptual components: $\textsc{Initialize}$, $\textsc{PlanShortestPath}$, and $\textsc{UpdateChange}$. The $\textsc{Initialize}$ procedure sets up the search structure with the object goal configuration fixed. $\textsc{PlanShortestPath}$ incrementally computes or repairs the shortest path from the current start configuration to the goal by updating only the states affected by previous changes. $\textsc{UpdateChange}$ handles environmental changes such as edge cost modifications, updating the affected edges, and propagating the changes to the planner. 
\begin{algorithm}[tb]
\vspace{0mm}
    \small
	\caption{$\rm \text{LAMP-}Lazy $}
\begin{algorithmic}[1]
\Require $\text{object } O, \text{robot }R, \text{policy } \text{GCo}_{DC}, \text{planner }\text{Gspi},\textsc{D*Lite}$
\State $\textsc{D*Lite}.\textsc{Initialize}(q_O^{goal})$
\State $\mathcal{T}, P_{O},Tree_{eval} \leftarrow \textsc{Plan}(q_O^{init}, q_R^{init}, \emptyset)$  \label{lazy:plan}
\While {$q_O^{now} \neq q_O^{goal}$}
\State $\mathcal{T}_R \leftarrow \mathcal{T}.pop(), q_O^{next} \leftarrow P_{O}.pop()$
\State $q_O^{now}, q_R^{now} \leftarrow \textsc{Execute}(\mathcal{T}_R)$\label{lazy:execute}
\If{$\textsc{Drift}(q_O^{now}, q_O^{next})$}\label{lazy:drift}
\State $\mathcal{T}, P_{O},Tree_{eval} \leftarrow \textsc{Plan}(q_O^{now}, q_R^{now}, Tree_{eval})$
\EndIf
\EndWhile
\Function{Plan}{$q_O^{now}, q_R^{now}, Tree_{eval}$}
\State $\mathcal{T} \leftarrow \emptyset; Tree_{new}\leftarrow\emptyset$\label{lazy:initialize}
\While{$\mathcal{T} = \emptyset$}
\State $P_{O} \leftarrow \textsc{D*Lite}.\textsc{PlanShortestPath}(q_O^{now})$\label{lazy:shortestpath}  
\State $q_O^{curr} \leftarrow q_O^{now}; q_R^{curr} \leftarrow q_R^{now}$
\For{$q_O^{next} \in P_{O}$}\label{lazy:examine-each-config}
\State $\textbf{if } q_O^{curr} \in Tree_{eval} {\rm \textbf{ then break}}$   \label{lazy:inthetree}
\State $m_O \leftarrow \textsc{GetPrimitive}(q_O^{curr},q_O^{next})$
\State $\mathcal{T}_{move},\mathcal{T}_{manip}, \mathcal{C}_{R} \leftarrow \textsc{Verify}(q_O^{curr},m_O,q_R^{curr})$\label{lazy:verify} 
\If{$\mathcal{C}_{R} = \emptyset$}
\State $\textsc{D*Lite}.\textsc{UpdateChange}(q_O^{curr},q_O^{next})$ \label{lazy:penalizeedge}
\State $\mathcal{T} \leftarrow \emptyset; Tree_{new}\leftarrow\emptyset$\label{lazy:discard}
\State $\textbf{break}$
\EndIf
\State $\mathcal{T}_R \leftarrow \mathcal{T}_{move} \oplus \mathcal{T}_{manip};\mathcal{T}.add(\mathcal{T}_R)$
\State $\mathcal{D}_{\mathcal{T}} \leftarrow \{q_R^{curr}: \mathcal{T}_R\}$
\State$Tree_{new}[q_O^{curr}] \leftarrow (q_O^{next},\mathcal{C}_{R}, \mathcal{T}_{manip}, \mathcal{D}_{\mathcal{T}})$ \label{lazy:treeupdate}
\State $q_O^{curr} \leftarrow q_O^{next};q_R^{curr} \leftarrow \mathcal{T}_{R}.last()$
\EndFor
\If{$q_O^{curr} \in Tree_{eval}$} \label{lazy:usetree}
\State $q_O^{next}, \mathcal{C}_{R}, \mathcal{T}_{manip}, \mathcal{D}_{\mathcal{T}} \leftarrow Tree_{eval}[q_O^{curr}]$\label{lazy:get-DT}
\If{$q_R^{curr} \notin \mathcal{D}_{\mathcal{T}}$}\label{lazy:not-in-DT}
\State $\textsc{Reassign}(\mathcal{C}_{R}, \mathcal{T}_{manip},q_R^{curr})$\label{lazy:reassign}
\State $\mathcal{T}_{move} \leftarrow \text{Gspi}(q_R^{curr},C_{R})$\label{lazy:Gspi}
\If{$\mathcal{T}_{move} = \emptyset$}
\State $\textsc{D*Lite}.\textsc{UpdateChange}(q_O^{curr},q_O^{next})$ 
\State $\mathcal{T} \leftarrow \emptyset; Tree_{new}\leftarrow\emptyset$
\State \textbf{break}
\EndIf
\State $\mathcal{T}_R \leftarrow \mathcal{T}_{move} \oplus \mathcal{T}_{manip}$\label{lazy:T_R}
\State $Tree_{eval}[q_O^{curr}].\mathcal{D}_\mathcal{T}[q_R^{curr}] \leftarrow \mathcal{T}_R$ \Comment{Cache}\label{lazy:cache}
\EndIf
\While{$q_O^{curr} \neq q_O^{goal}$}\label{lazy:backtrack-while}
\State $q_O^{next}, \mathcal{C}_{R}, \mathcal{T}_{manip}, \mathcal{D}_{\mathcal{T}} \leftarrow Tree_{eval}[q_O^{curr}]$
\State $\mathcal{T}_R \leftarrow \mathcal{D}_{\mathcal{T}}[q_R^{curr}]$ \Comment{Reuse cached}
\State $\mathcal{T}.add(\mathcal{T}_R)$
\State $q_O^{curr} \leftarrow q_O^{next};q_R^{curr} \leftarrow \mathcal{T}_{R}.last()$ \label{lazy:usetreeend}
\EndWhile
\EndIf 
\EndWhile  
\State $Tree_{eval} \leftarrow Tree_{new} \cup Tree_{eval}$\label{lazy:update-tree-eval}
\State \Return $\mathcal{T}, P_{O}, Tree_{eval}$\label{lazy:return}
\EndFunction
\end{algorithmic}
\label{alg:Lazy}
\vspace{-0.1cm}
\end{algorithm}

We adopt D* Lite as an object-centric planner and defer verification until an object path is returned. We constantly iterate over this plan-verify-execute loop. When verification of an object transition fails, we penalize the corresponding edge by increasing its cost, encouraging D* Lite to explore alternative paths in subsequent iterations. The detailed procedure of our approach is shown in Alg.~\ref{alg:Lazy}. We first initialize the D* Lite planner and invoke $\textsc{Plan}$ (line~\ref{lazy:plan}) to obtain the initial robot trajectories $\mathcal{T}$, the corresponding object path $P_{O}$, and a globally maintained evaluation tree $Tree_{eval}$ that caches validated transitions together with their manipulation trajectories, which will later be elaborated when introducing the $\textsc{Plan}$ function. We then execute each segment of $\mathcal{T}$ in sequence (line~\ref{lazy:execute}) and observe the resulting positions of the robots and the object. If the observed drift of the object exceeds a pre-defined threshold (line~\ref{lazy:drift}), we immediately replan from the current configuration to obtain an updated plan. This process is fast due to the reuse of planning efforts, and repeats until the object reaches its goal configuration. 

We now elaborate on the core function $\textsc{Plan}$. It generates executable manipulation trajectories $\mathcal{T}$ to transport the object to the goal configuration by combining incremental path planning with cached validation results. The key data structure is the evaluation tree $Tree_{eval}$, which stores previously validated object transitions from $q_O^{curr}$ to $q_O^{next}$  along with their contact points $\mathcal{C}_R$ and manipulation trajectories $\mathcal{T}_{manip}$, enabling efficient reuse of prior planning effort. Specifically, each entry is represented as $Tree_{eval}[q_O^{curr}] = (q_O^{next},\mathcal{C}_{R}, \mathcal{T}_{manip}, \mathcal{D}_{\mathcal{T}})$, where $\mathcal{D}_{\mathcal{T}}$ is a cache that maps different robot configurations to full trajectories, as described later. Importantly, we store only transitions that lie on paths that move an object from a configuration to the goal configuration $q_O^{goal}$ in  $Tree_{eval}$, forming a tree rooted at $q_O^{goal}$.

We use $\mathcal{T}$ to store the full-horizon path and $Tree_{new}$ to store the newly validated transitions in each iteration (line~\ref{lazy:initialize}). We first invoke D* Lite to compute an object-centric path $P_O$ from the current configuration to the goal (line~\ref{lazy:shortestpath}) and then verify the feasibility of each consecutive configuration pair in $P_O$. For each pair $(q_O^{curr}, q_O^{next})$ (line~\ref{lazy:examine-each-config}), if the current object configuration $q_O^{curr}$ already exists in $Tree_{eval}$, we stop the verification and reuse the cached transitions (line~\ref{lazy:inthetree}). Otherwise, we call $\textsc{Verify}$ (line~\ref{lazy:verify}), as in LAMP-A*, to validate the transition, yielding motion trajectories $\mathcal{T}_{move}$, manipulation trajectories $\mathcal{T}_{manip}$, and contact points $\mathcal{C}_R$. If verification fails, we penalize the corresponding edge by increasing its cost (line~\ref{lazy:penalizeedge}), discard all transitions in $Tree_{new}$ to prevent backtracking along an infeasible partial path (line~\ref{lazy:discard}), and re-call D* Lite to plan a new object-centric path $P_O$ based on the updated cost (line~\ref{lazy:shortestpath}). Otherwise, we construct the complete trajectory $\mathcal{T}_R$, add it to $\mathcal{T}$, and initialize a cache $\mathcal{D}_\mathcal{T}$ mapping the current robot configurations to $\mathcal{T}_R$. 
These validated results are temporarily stored in $Tree_{new}$ for later insertion into $Tree_{eval}$ (line~\ref{lazy:treeupdate}). We then advance both the object and robot configurations to the next configuration in $P_O$ and continue verification. 

If the entire path is verified without reaching any configuration in $Tree_{eval}$, the accumulated trajectories in $\mathcal{T}$, as well as the cached verified transitions in $Tree_{new}$, already form a full-horizon plan to the goal which can be directly committed to $Tree_{eval}$ (line~\ref{lazy:update-tree-eval}) and return (line~\ref{lazy:return}). Otherwise, when verification reaches a configuration in $Tree_{eval}$, we perform backtracking (lines~\ref{lazy:usetree}--\ref{lazy:usetreeend}) to assemble the remaining portion of the manipulation plan. One caveat is that, while the current object configuration $q_O^{curr}$ is in $Tree_{eval}$, the current robot configuration $q_R^{curr}$ may not be found in the corresponding $\mathcal{D_T}$ (lines~\ref{lazy:get-DT}--\ref{lazy:not-in-DT}), which means that we do not yet have robot motion trajectories to the contact points. In such a case, we compute $T_{move}$ as in the \textsc{Verify} (lines~\ref{lazy:reassign}--\ref{lazy:Gspi}). If no valid $T_{move}$ is found, we penalize the corresponding edge in D* Lite and go to next iteration to re-plan a new object-centric path. Otherwise, we create the full trajectory $\mathcal{T}_R$ (line~\ref{lazy:T_R}) and cache it in the corresponding $\mathcal{D_T}$ (line~\ref{lazy:cache}). 

Then, starting from the object configuration where verification stopped, we iteratively trace $Tree_{eval}$ toward the goal, retrieving at each step the next object configuration, its associated manipulation trajectories, and the trajectory cache $\mathcal{D}_{\mathcal{T}}$ (lines~\ref{lazy:backtrack-while}--\ref{lazy:usetreeend}). 
Upon successful construction of the full-horizon manipulation trajectories, we commit all newly validated transitions from $Tree_{new}$ to $Tree_{eval}$ and return the full-horizon plan $\mathcal{T}$ for execution (lines~\ref{lazy:update-tree-eval}--\ref{lazy:return}). 
\vspace{-0.05cm}

\section{Experiments}
\label{sec:experiments}
We evaluate our approach on challenging simulated scenarios that require solving complex long-horizon tasks in highly cluttered environments. Our experiments compare against state-of-the-art baselines across diverse scenarios to assess both solution quality and computational efficiency.
\begin{figure}[t]
\vspace{0.17cm}
\centering
\includegraphics[width=0.6\columnwidth]{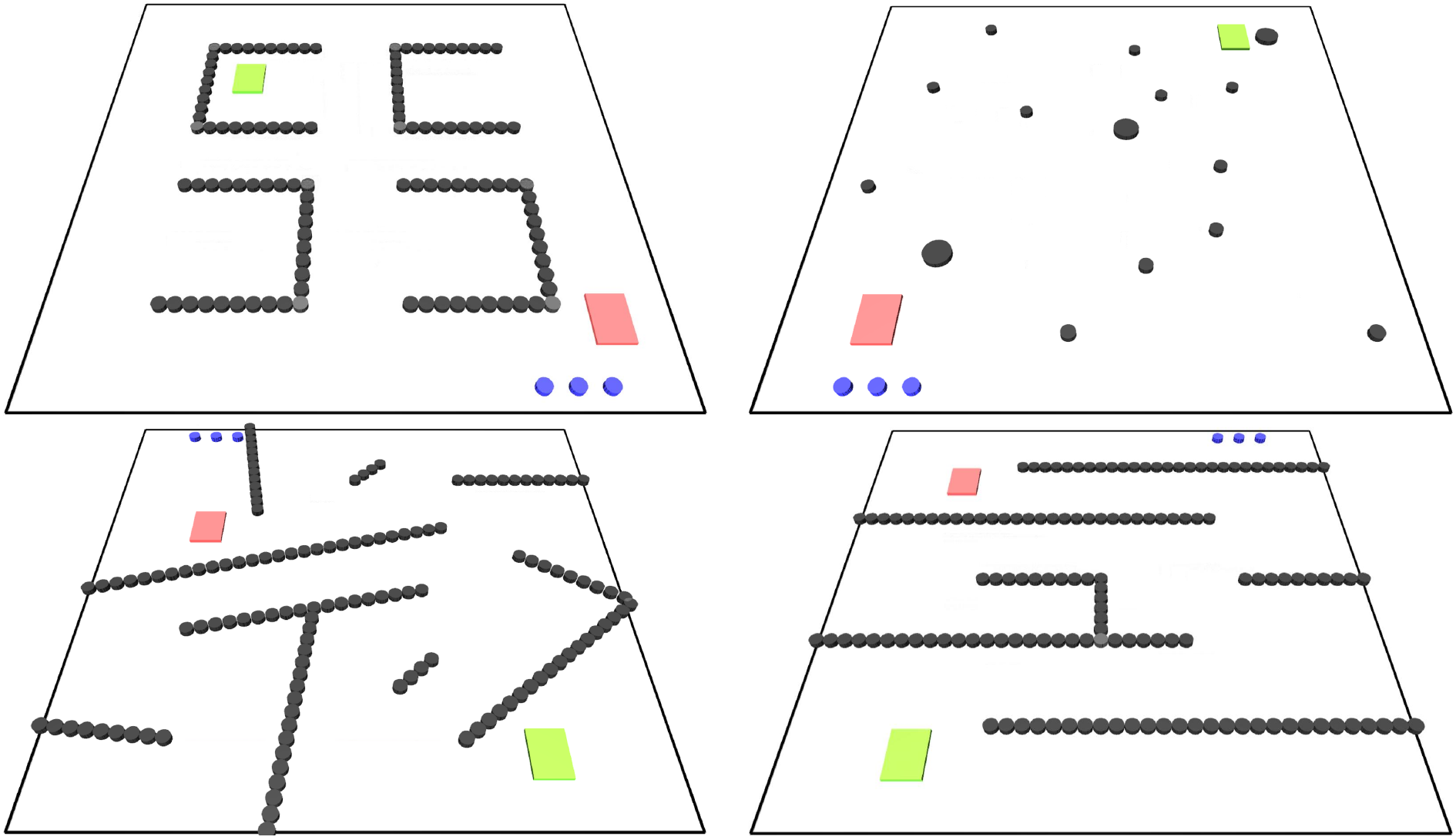}
\caption{Test cluttered maps with obstacles (black), robots (blue), example object start positions (red), and goal positions (green). From top to bottom and left to right: Warehouse, Random, Tilt, and Maze.}
\label{fig:map}
\vspace{-0.67cm}
\end{figure}

\subsection{Single-Object Manipulation Experiments}

We conducted experiments on four distinct maps (Random, Maze, Tilt, Warehouse) (Fig.~\ref{fig:map}), each with different obstacle configurations, workspace layouts, and clutter densities. Specifically, the Random map is relatively sparse while the other three maps are more cluttered. For each map, we designed 25 different scenes with varying object start and goal configurations, object types (circles and rectangles), and object sizes. Each scene requires a team of three robots to transport an object to its goal configuration through collaborative pushing, navigating through narrow passages between obstacles. This creates varying levels of difficulty depending on the object size, obstacle placement, and available free space, yielding a total of 100 test scenarios across all maps.

Since we do not assume access to contact models or specific physical parameters, we compare against following learning-based and hybrid methods: 
\textbf{MAPush}~\cite{fyuming2025MAPush}, a hierarchical reinforcement learning approach that decouples low-level motion control from high-level robot--object interaction strategies;\footnote{We used the open-source implementation (github.com/collaborative-mapush/MAPush) with 3 robots and trained from scratch for 100M steps using the original MQE framework and reward structure.} 
\textbf{GCo}~\cite{shaoul2025Gco}, a state-of-the-art baseline combining learned manipulation with planning introduced in Sec. \ref{bg:gco} (we evaluate both the original formulation $\text{GCo}_{ori}$ with conservative safety buffer inflation around obstacles, and a modified variant $\text{GCo}_{var}$ with zero buffer constraints to enable operation in tighter spaces);\footnote{We used the open-source implementation (github.com/yoraish/gco) and trained $\text{GCo}_{\text{DC}}$ with the original 20,000-sample dataset collected in MuJoCo~\cite{todorov2012mujoco}. This model was also used by our LAMP-A* and LAMP-Lazy.} 
\textbf{LAMP-A* (ours)}, the planner as described in Alg.~\ref{alg:Astar} (we evaluate two variants: open-loop execution $\text{LAMP-}A^*_{ori}$ and closed-loop execution with replanning triggered upon drift detection $\text{LAMP-}A^*_{replan}$); 
and \textbf{LAMP-Lazy (ours)}, our D* Lite-based planner with closed-loop replanning, as described in Alg.~\ref{alg:Lazy}.

We allow these methods generous computation limits to focus on manipulation capacity: up to 100 execution iterations of manipulation for GCo and our LAMP and up to 2 simulated minutes for MAPush. Additionally, for LAMP-A*, we impose a cumulative planning time limit of 500s to prevent excessive computation during search. A scenario is considered successful if it satisfies the method-specific\footnote{Following~\cite{shaoul2025Gco}, MAPush is successful if the object reaches within 0.5\,m of the goal position while ignoring orientation. For all other methods, success requires the final object pose to be within 0.1\,m of the goal position in translation and within 0.1\,rad in orientation.} success criterion above within these execution and time budgets while keeping the object entirely inside the map throughout execution; otherwise, it is marked as a failure.
\begin{table}[t]
\vspace{0.11cm}
\centering
\caption{Overall success rates across different maps.}
\label{tab:success_rates}
\begin{tabular}{lcccc}
\toprule
\textbf{Method} & \textbf{Random} & \textbf{Maze} & \textbf{Tilt} & \textbf{Warehouse} \\
\midrule
MAPush & 8\% & 0\% & 0\% & 0\% \\
$\text{GCo}_{ori}$ & 0\% & 0\% & 0\% & 0\% \\
$\text{GCo}_{var}$ & 68\% & 12\% & 28\% & 20\% \\
$\text{LAMP-}A^*_{ori}$ (ours) & 24\% & 0\% & 0\% & 12\% \\
$\text{LAMP-}A^*_{replan}$ (ours) & 88\% & 0\% & 8\% & 0\% \\
\textbf{LAMP-Lazy (ours)} & \textbf{100\%}  & \textbf{96\%} & \textbf{100\%} & \textbf{88\%} \\
\bottomrule
\end{tabular}

\vspace{0.5em}

\caption{Full-horizon planning success rate (complete paths found before execution).}
\label{tab:full_horizon_rate}
\begin{tabular}{lcccc}
\toprule
\textbf{Method} & \textbf{Random} & \textbf{Maze} & \textbf{Tilt} & \textbf{Warehouse} \\
\midrule
$\text{LAMP-A*}$ & 100\% & 8\% & 32\% & 40\% \\
\textbf{LAMP-Lazy} & \textbf{100\%} & \textbf{100\%} & \textbf{100\%} & \textbf{100\%} \\
\bottomrule
\end{tabular}
\vspace{-0.6cm}
\end{table}

\begin{figure}[t]
\vspace{0.075cm}
\centering
\includegraphics[width=.85\columnwidth]{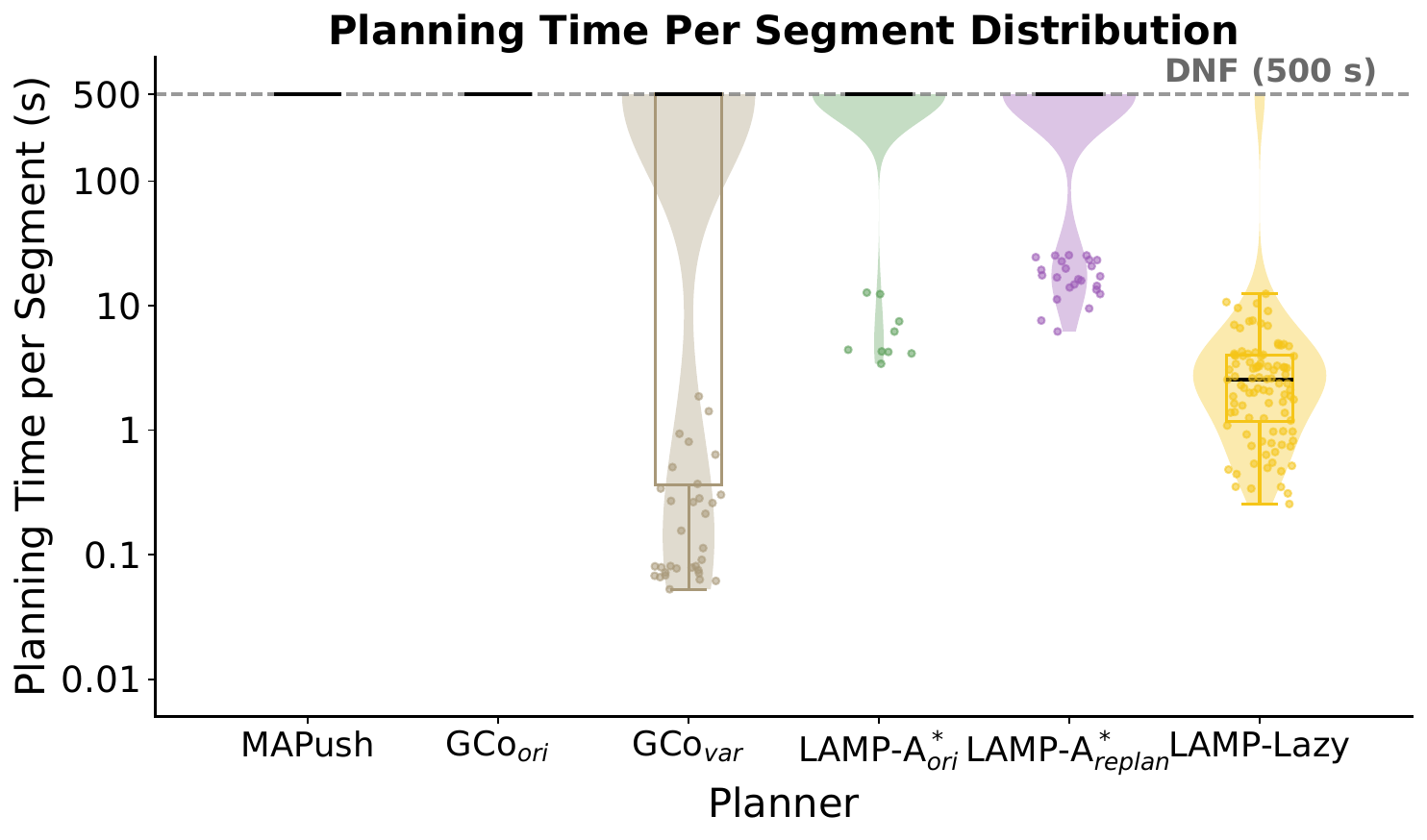}
\caption{Planning time per segment across 100 scenarios. Each violin plot shows the distribution shape, with individual dots representing scenario values. The box inside indicates median and interquartile range. Failed cases are assigned 500s and shown at the dashed line.}
\label{fig:planning_time}
\vspace{-0.5cm}
\end{figure}

\begin{figure}[t]
\tiny
\centering
\includegraphics[width=0.7\columnwidth]{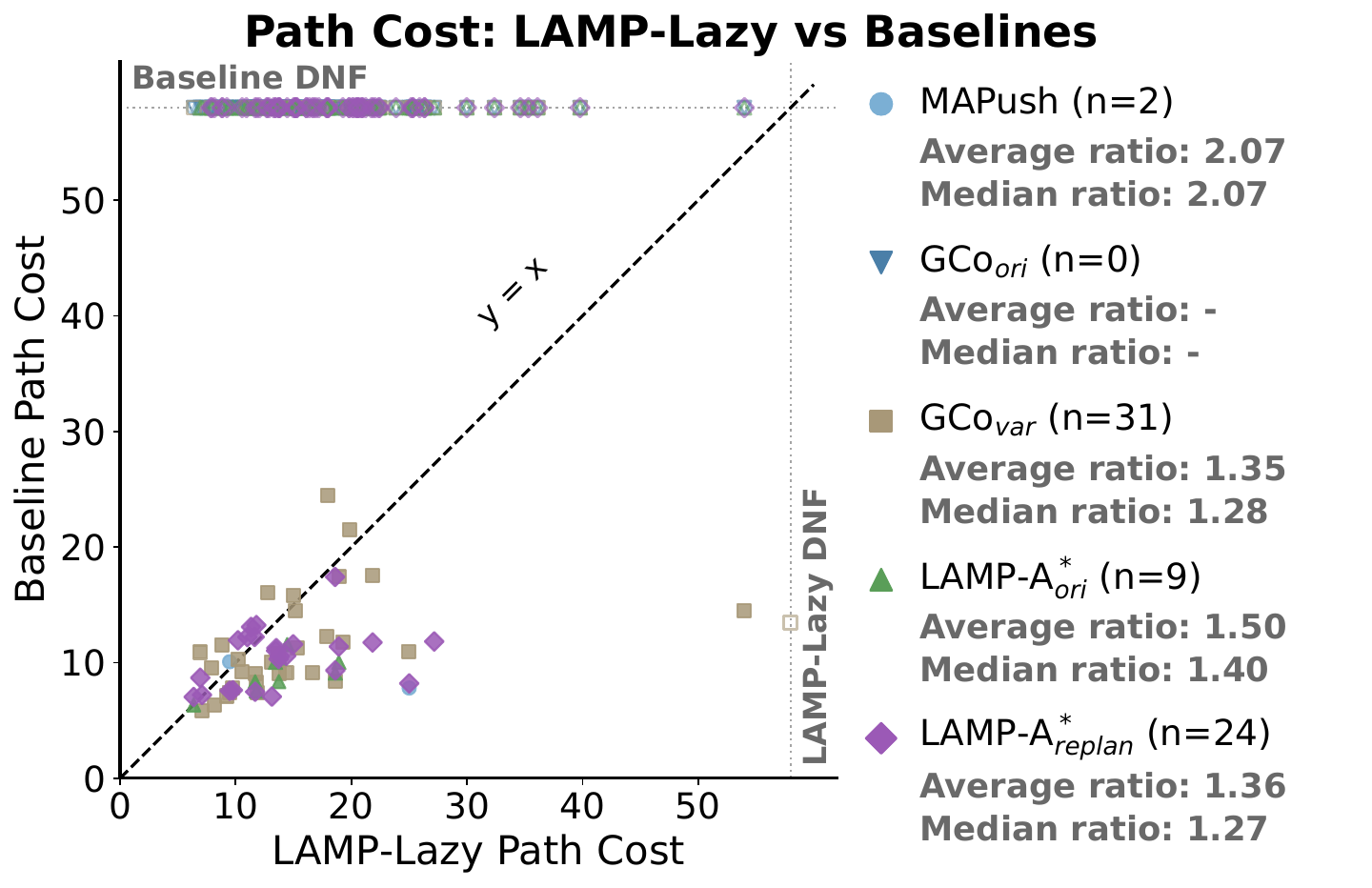}
\caption{Path cost across 100 scenarios. Each point represents a case solved by both LAMP-Lazy and the corresponding baseline. In the legend, n denotes the number of such cases, and ratios are computed as the path cost of LAMP-Lazy divided by that of the corresponding baseline over these cases.  DNF (Did Not Finish) indicates one-sided failures and is excluded from the ratio statistics.}
\label{fig:path_cost}
\vspace{-0.65cm}
\end{figure}

\subsubsection{Success rate} 

Table~\ref{tab:success_rates} shows the percentage of scenarios where the object reached the goal within execution and time budgets. LAMP-Lazy achieves the highest success rate across all environments. The few failures of LAMP-Lazy occur in Maze and Warehouse. In these cases, the planned object paths are close to workspace boundaries, and the push trajectories proposed by $\text{GCo}_{DC}$ cause the object to intersect the workspace boundary during execution, which is less likely in the more open Random map.
 Among baselines, $\text{GCo}_{var}$ performs second-best, succeeding in sparse scenarios (Random) but degrading in cluttered maps where robots become trapped in corners due to lack of robot-level feasibility verification. $\text{LAMP-}A^*_{replan}$ finds many full-horizon paths (Table~\ref{tab:full_horizon_rate}) but requires excessive planning time, failing to reach goals within our 500s limit. $\text{LAMP-}A^*_{ori}$ suffers from both long planning times and execution divergence. MAPush and $\text{GCo}_{ori}$ fail in most scenarios due to obstacle-agnostic navigation and overly conservative safety buffers.

\subsubsection{Planning time per segment}
Fig.~\ref{fig:planning_time} shows the planning time per manipulation segment across all scenarios. LAMP-Lazy achieves substantially lower planning time than both $\text{LAMP-}A^*_{ori}$ and $\text{LAMP-}A^*_{replan}$, demonstrating the efficiency gains from lazy evaluation. The replanning variant $\text{LAMP-}A^*_{replan}$ requires notably more computation than $\text{LAMP-}A^*_{ori}$ due to repeated search invocations during execution. $\text{GCo}_{var}$ exhibits faster per-segment times, as expected for a method that directly executes contact-based primitives after model inference without explicit search. MAPush and $\text{GCo}_{ori}$ show insufficient data points due to their low success rates. Overall, LAMP-Lazy maintains planning time within a practical range (median approximately 2--3 seconds per segment) while achieving high success rates.

\subsubsection{Path cost}
We compare the makespan of robot manipulation trajectories across all scenarios in Fig.~\ref{fig:path_cost}. The plot reveals that LAMP-Lazy generally produces solutions with comparable or slightly higher path costs compared to baselines in mutually solvable scenarios. Among the baselines, $\text{GCo}_{var}$ (brown squares) and $\text{LAMP-}A^*_{replan}$ (purple diamonds) tend to find lower-cost solutions when they succeed. This is expected, as LAMP-Lazy prioritizes planning efficiency and success rate over path optimality. Nevertheless, the path cost difference remains modest in most scenarios, demonstrating that LAMP-Lazy significantly improves success rates without substantial degradation in solution quality.
\begin{figure}[t]
\vspace{0.18cm}
\centering
\includegraphics[width=.6\columnwidth]{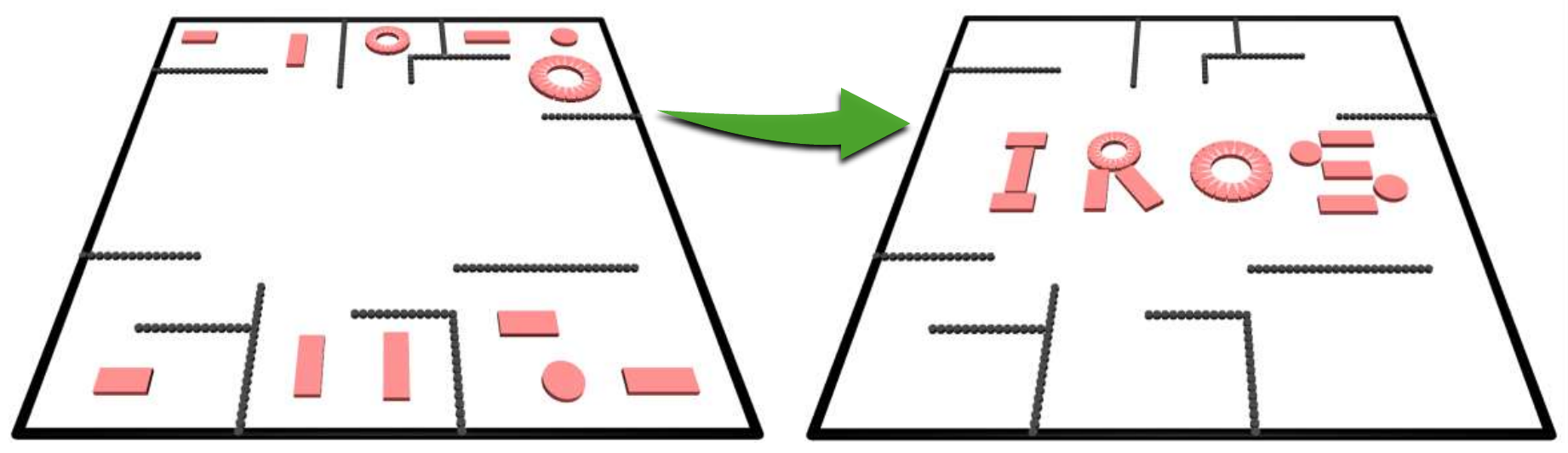}
\caption{Long-horizon manipulation task: assembling ``IROS'' with 9 objects. See supplemental video.}
\label{fig:iros}
\vspace{-0.65cm}
\end{figure}
\vspace{-0.5cm}
\subsection{Long-Horizon Sequential Manipulation Case Study}

To evaluate scalability and robustness over extended sequences, we include a qualitative long-horizon stress test in which robots transport 9 objects one-by-one to assemble the ``IROS'' logo in a warehouse-like environment (Fig.~\ref{fig:iros}). As objects are progressively placed, the workspace becomes increasingly cluttered, compounding the difficulty for later objects. 
We tested all methods under the same constraints. MAPush and $\text{GCo}_{var}$ failed at various stages when robots became trapped in corners during manipulation, unable to reposition effectively. $\text{LAMP-}A^*_{replan}$ and $\text{LAMP-}A^*_{ori}$ failed due to exceeding the 500s planning time limit as clutter increased. LAMP-Lazy successfully transported all 9 objects to their goals with an average planning time of only 2.07 seconds per segment, demonstrating robust performance on challenging long-horizon sequential tasks.

\section{Conclusions}
\label{sec:conclusions}
	We presented LAMP, a unified framework for long-horizon multi-robot manipulation planning that integrates manipulation feasibility verification directly into object-centric search. We introduced LAMP-A*, which systematically explores the joint space of object and robot configurations with eager validation, and LAMP-Lazy, which enables efficient replanning through lazy evaluation. Experiments across diverse cluttered environments demonstrate that LAMP-Lazy substantially outperforms prior methods in success rate while maintaining practical planning times. Although LAMP-Lazy produces slightly higher-cost solutions compared to baselines on mutually solvable scenarios, this modest trade-off enables robust closed-loop execution on challenging long-horizon sequential manipulation tasks where prior approaches fail.

\section*{Acknowledgments}
This work was partially supported by the National Science Foundation under grant numbers \#$2328671$ and \#$2441629$. 

\addtolength{\textheight}{-12cm}   




\bibliographystyle{IEEEtran}
\bibliography{references}

\end{document}